\documentclass{article}

\usepackage[english]{babel}

\usepackage[letterpaper,top=2cm,bottom=2cm,left=3cm,right=3cm,marginparwidth=1.75cm]{geometry}

\usepackage{amsmath}
\usepackage{amsthm}
\usepackage{amsfonts}
\usepackage{graphicx}
\usepackage{mathtools}
\usepackage[colorlinks=true, allcolors=blue]{hyperref}
\usepackage{tikz,pgfplots}
\usepackage[square, numbers, comma, sort&compress]{natbib}
\usepackage{dirtytalk}
\usepackage{caption}
\usepackage{subcaption}
\usetikzlibrary{shapes.geometric,arrows}
\usetikzlibrary{bayesnet}
\usetikzlibrary{backgrounds}
\usetikzlibrary{matrix,calc,patterns,decorations.pathmorphing,decorations.markings}

\theoremstyle{definition}
\newtheorem{definition}{Definition}[section]
\newtheorem{theorem}{Theorem}

\title{On the use of Statistical Learning Theory for model selection in Structural Health Monitoring}
\author{C. A. Lindley, N. Dervilis, K. Worden}

\begin{document}
\maketitle

\begin{abstract}
    Whenever data-based systems are employed in engineering applications, defining an optimal statistical representation is subject to the problem of model selection. This paper focusses on how well models can generalise in Structural Health Monitoring (SHM). Although statistical model validation in this field is often performed heuristically, it is possible to estimate generalisation more rigorously using the bounds provided by Statistical Learning Theory (SLT). Therefore, this paper explores the selection process of a kernel smoother for modelling the impulse response of a linear oscillator from the perspective of SLT. It is demonstrated that incorporating domain knowledge into the regression problem yields a lower guaranteed risk, thereby enhancing generalisation.
\end{abstract}

\section{Introduction}

The incorporation of intelligent systems in engineering is a subject of increasing popularity in the literature. Of particular interest in the present study is the development of data-based methods tailored to addressing various challenges encountered in \textit{Structural Health Monitoring} (SHM)~\cite{Farrar2013}. Whether these methods are designed to detect, localise or classify damage~\cite{Rytter1993}, one aspect that is shared across them is the problem of model selection; namely, of deciding over a set of statistical models, which one can adequately represent the system at hand. The means to determine such a model is by a process of validation during the training phase, which is a crucial step to ensure that the model can genenralise well. Failure to do so can lead to an erroneous representation of the system, and thus inaccurate predictions upon the introduction of new observations. This consideration may arguably be of greater importance in SHM, since poor predictions could be perilous for human safety, and detrimental towards financial projections, in the event of unforeseen failures.

To ensure good generalisation, standard practice is followed, often involving the division of the entire dataset into a \textit{training set} and a \textit{test set}. In short, the training set is used to search for optimal parameters, while the test set simulates unseen data, allowing the evaluation of the prediction error for both sets during the learning process. Some optimal model is chosen that can maintain a low error for the training set while preventing the error to grow large for the test set. This criterion is commonly met heuristically with the use of \textit{cross-validation} methods~\cite{Bishop2006}. The concept of generalisation, however, can be approached in a more rigorous fashion and thus assert the confidence one has about the employed statistical model.

The search for an optimal model is thereby explored here by eliciting the methods found in \textit{Statistical Learning Theory} (SLT)~\cite{Vapnik1999}; concretely, in the selection of an optimal kernel smoother for modelling the impulse response of a simple structure. The premise of the following study is to establish a rigorous mathematical framework for the model-selection problem in SHM, ensuring the selection of an optimal model that generalises well when trained with a limited amount of data. 

\section{The model-selection problem}

To illustrate the model-selection problem, one may first consider the case in which the aim is to fit a function to a set of data. If dealing with a regression problem, then a sensible approach would be to find some weighted combination of inputs that yields values closely approximating their corresponding targets. In other words, to develop a function $f$ that can accurately map a $D$-dimensional input vector $\textbf{x}_n \in \mathbb{R}^D$, to a target output scalar $y_n \in \mathbb{R}$, i.e.\ a predictor $f:\textbf{x} \to y$. The fit on $y$ can be thought of as an interpolation given by $f$.

In a statistical sense, the predictor is a function learnt from the data, albeit conditioned on certain assumptions. For example, a linear interpolant may be assumed to fit the data, which will likely result in a poor fit if the true underlying trend is strongly nonlinear. An improved fit can be achieved by increasing the order of the function; that is, a quadratic function will likely provide a better fit than the linear trend, a cubic will be better than the quadratic, and so on. At each step, the fit improves while the function becomes increasingly complex, requiring the addition of more parameters to account for the higher-order terms in the polynomial. Continuing with the inclusion of terms, however, can lead to the point where the function is said to have become too complex, and the model begins to (over)fit noise in the data. Somewhere in this process of adding terms to the predictor, an optimal amount of complexity is attained, whereby the complexity of the data is matched.

Whether the learning problem is of regression or classification, finding some optimal predictor is typically approached by minimising a measure of discrepancy (or \textit{loss}) between the true target values and the corresponding outputs of the predictor. To illustrate, let $\mathrm{z} = (\mathrm{x},y)$ denote an input-output pair in the training set, and $Q(\mathrm{z},\theta)$, $\theta\in\Theta$ be the set of loss functions. When given a set of $n$ i.i.d.\ samples $\mathrm{z}_1,\dots,\mathrm{z}_n$, the goal of predictive learning is to find a function $Q(\mathrm{z},\theta^*)$ that minimises the \textit{expected risk},
\begin{equation}
    R(\theta) = \int \! Q(\mathrm{z},\theta) p(\mathrm{z}) \mathrm{d}\mathrm{z}
    \label{eq:expected_risk}
\end{equation}
where $Q(\mathrm{z},\theta) = L(\mathrm{y},f(\mathrm{x},\theta))$ is some \textit{loss function}, and the integral is evaluated with respect to some (unknown) joint probability distribution $p(\mathrm{z})$. One should note that the set $\Theta$ to which $\theta$ belongs can be a set of scalar quantities, vectors, or of abstract elements~\cite{Vapnik2006}; that is, the function space is not limited to parametric models.

The current framework assumes that the joint distribution is true but unknown, and that the only information that is made available is samples drawn from $p(\mathrm{z})$. In order to minimise the risk functional $R(\theta)$, the following inductive principle is considered,
\begin{equation}
    R_{emp}(\theta) = \frac{1}{n} \sum_{i=1}^n Q(\mathrm{z_i},\theta)
    \label{eq:empirical_risk}
\end{equation}
where $R_{emp}(\theta)$ is referred to as the \textit{empirical risk}, which is evaluated with a finite sample of size $n$. Given that the empirical risk does not require knowing the underlying generative distribution, the idea is to seek for an estimate providing the minimum empirical risk~(\ref{eq:empirical_risk}), in hopes that such estimate will also minimise the true risk~(\ref{eq:expected_risk}). This approach is called the \textit{Empirical Risk Minimisation inductive principle} (ERM principle)~\cite{Vapnik1999}.

While minimising the empirical risk~(\ref{eq:empirical_risk}) promotes learning, a compromise must still be maintained by having a small enough risk without producing an overly complex model. This trade-off is crucial to ensuring good generalisation. A possible solution to this issue is to estimate the expected risk as a function of the empirical risk, penalised by some measure of model complexity~\cite{Cherkassky2007}; that is,
\begin{equation}
    R(\theta) \cong r \left ( \frac{h}{n} \right ) R_{emp}
    \label{eq:penalised_risk}
\end{equation}
where the empirical risk is adjusted by a monotonically-increasing function $r$, defined by the ratio of some \textit{capacity measure} $h$, over the sample size $n$ \cite{Hardle1988}.

It turns out that the criteria for defining the penalisation function arises naturally in SLT. The foundation of such a criteria derives from demonstrating, in a rigorous mathematical framework, that minimising the empirical risk~(\ref{eq:empirical_risk}) can in fact yield a small value of the actual risk~(\ref{eq:expected_risk}). In short, if one can show that the empirical risk \textit{converges uniformly} to the true expected risk, then one can be (almost) certain that the ERM principle leads to generalised models. This condition is presented in more detail by the following definition:

\begin{definition}[Key Theorem of Learning Theory~\cite{Vapnik1999}]
    \textit{For a finite set of bounded loss functions, the ERM inductive principle is consistent if, and only if, the empirical risk \textit{converges uniformly} to the true risk,}
    \begin{equation}
        \lim_{n\to\infty} P \left \{ \sup_{\theta\in\Theta} \left | R(\theta) - R_{emp}(\theta) \right | > \epsilon \right \} = 0, \quad \forall \epsilon > 0
        \label{eq:key_theorem}
    \end{equation}
\label{def:uniform_consistency}
\end{definition}

In the light of this theorem, a bound can be determined on the expected risk, whereby the penalised function is defined. In particular, for regression problems, the implementation of the ERM inductive principle defines the generalisation ability of a learning machine as demonstrated by the following result:
\begin{theorem}
    (Vapnik-Chervonenkis~\cite{Vapnik2006})\textit{ For any $\delta\in(0,1)$, with probability at least $(1-\delta)$ and $\forall \theta \in \Theta$,}
    \begin{equation}
        R(\theta) \leq \frac{R_{emp}(\theta)}{(1-c\sqrt{\eta})_{+}}
        \label{eq:regression_bound}
    \end{equation}
    \textit{where}
    \begin{equation}
        \eta = \eta \left ( \frac{n}{h},\frac{-\ln \delta}{n} \right ) = a_1 \frac{h[\ln(a_2n/h) + 1] - \ln (\delta/4)}{n}
    \end{equation}
    \textit{when the set of loss functions} $Q(\mathrm{z},\theta), \theta \in \Theta$ \textit{is nonnegative, unbounded and contains an infinite number of elements. The measure $h$ quantifies the capacity of the function class, and the constants $a_1$, $a_2$ and $c$ are adjusted to suit the type of learning problem.}
    \label{th:regression_bound}
    \end{theorem}

Deriving this bound is, indeed, an involved process, and the final result is presented here without proof. However, a full derivation can be found in~\cite{Vapnik2006}. In the interest of the current work, this bound forms the basis for the development of the model-selection strategy followed in the case study below.

Before delving into the uses of the presented theory in SHM, an additional inductive principle must be reviewed. This principle corresponds to the \textit{Structural Risk Minimisation} also formalised by Vapnik and Chervonenkis~\cite{Vapnik1999}. It is shown next how SRM sets the stage for a systematic model-selection criteria. Additionally, the concept of complexity is further clarified.

\section{Managing complexity and minimisation of structural risk}

So far, complexity has been implicitly defined by an increasing addition of terms - and thus parameters - in the predictor. At first, this definition seems to make sense, but the definition of complexity in SLT is more elaborate than merely taking the count of parameters in the model. In (\ref{eq:regression_bound}), the \textit{capacity} measure aims to rigorously quantify such complexity. Specifically, $h$ corresponds to a specific type of capacity measure known as the \textit{VC dimension}~\cite{Vapnik1999} (abbreviation for Vapnik and Chervonenkis dimension). The VC dimension is formally defined as follows.
\begin{definition}[VC dimension~\cite{Vapnik1999}]
    \textit{Let $S_{\Theta}(n)$ be the number of ways into which a sample of size $n$ can be classified by the function class $\Theta$. Therefore, the VC dimension of a function class $\Theta$ is the largest sample size $n$ such that the condition}
    \begin{equation}
        S_{\Theta}(n) = 2^n
    \end{equation}
    \textit{is valid.}
\end{definition}
In other words, the VC dimension $h$, corresponds to the size of the largest sample that a given set of indicator functions can \textit{shatter}\footnote{A function space is said to shatter the observations when a sample of size $n$ exists on which all possible dichotomies can be achieved by the set of indicator functions.}. If the set of functions under consideration can generate any classification on the sample, regardless of $n$, then the corresponding VC dimension is infinite. Having an infinite VC dimension translates into having a learner capable of fitting any collection of observations, and thus no valid generalisation is possible; that is, a function in the space exists such that any set of data can be explained, or equivalently, \textit{overfitted}.

Many important generalisation bounds in SLT make use of the VC dimension, included the one presented in Theorem \ref{th:regression_bound}. Estimating the VC dimension for different sets of function is, therefore, an essential step in the implementation of error bounds in SLT. However, analytic estimates are often only be determined for simple sets of functions. For example, a set of parametric linear estimators has a fixed complexity (VC dimension) that is equal to the number of free parameters. It is important to note that this is a coincidence and that the estimation of the VC dimension is not always this straightforward.

Once the complexity is defined for given set of functions, estimates of the expected risk no longer depend on the underlying distribution from which the sample is assumed to have been drawn. This implication is advantageous since the problem of density estimation is ill-posed when no prior knowledge of the probability distribution is available~\cite{Vapnik2006}.

Now, recalling the general problem of model selection, one can leverage on the complexity of the function space to choose an optimal solution for a finite sample. This premise is the basis of the SRM inductive principle for minimisation of the expected risk. Essentially, SRM operates by establishing a structure on some set $S$ of loss functions $Q(\mathbf{z},\theta)$, $\theta \in \Theta$, consisting of nested subsets (or elements) $S_k = \{Q(\mathbf{z},\theta)$, $\theta \in \Omega_k$\} such that,
\begin{equation}
    S_1 \subset S_2 \subset \dots \subset S_k \subset \dots
    \label{eq:fun_struct}
\end{equation}
where each element in the structure has a finite VC dimension $h_k$, and can be ordered according to their respective complexities,
\begin{equation}
    h_1 \leq h_2 \leq \dots \leq h_k \leq \dots
    \label{eq:vc_struct}
\end{equation}
Having defined the structure $S$, the optimal model selection boils down to two steps:
\begin{enumerate}
    \item Selecting an element $S_k$ from the structure.
    \item Estimating the model from this element.
\end{enumerate}

For each element in the structure (step ($1$)), the bound~\eqref{eq:regression_bound} is evaluated after estimating the optimal parameters in the element that minimise the empirical risk (step ($2$)). The penalisation during the learning process is thus established by the denominator in~\eqref{eq:regression_bound}, which is computed with respect to the VC dimension $h_k$ corresponding to the element $S_k$. Therefore, as elements of higher complexity are selected, the empirical risk will likely diminish, but at expense of a higher penalisation as a result of bringing the denominator closer to zero. An optimal element in the structure will be that providing the minimal \say{guaranteed} risk.

\section{Case study: Model selection for modelling a SDOF impulse response}

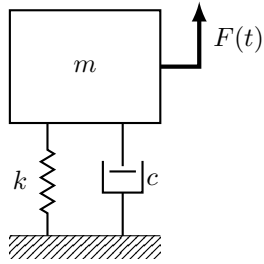
\begin{figure}[t]
    \hspace*{6cm}\begin{tikzpicture}[every node/.style={draw,outer sep=0pt,thick}]
    
        \tikzstyle{spring}=[thick,decorate,decoration={zigzag,pre length=0.3cm,post length=0.3cm,segment length=6}]
        \tikzstyle{damper}=[thick,decoration={markings,  
            mark connection node=dmp,
            mark=at position 0.5 with 
            {
            \node (dmp) [thick,inner sep=0pt,transform shape,rotate=-90,minimum width=15pt,minimum height=10pt,draw=none] {};
            \draw [thick] ($(dmp.north east)+(2pt,0)$) -- (dmp.south east) -- (dmp.south west) -- ($(dmp.north west)+(2pt,0)$);
            \draw [thick] ($(dmp.north)+(-2pt,-5pt)$) -- ($(dmp.north)+(-2pt,5pt)$);
            }
        }, decorate]
        \tikzstyle{ground}=[fill,pattern=north east lines,draw=none,minimum width=1cm,minimum height=0.3cm]
    
        \node (M) [minimum width=2.0cm,minimum height=1.5cm] {$m$};
    
        \node (ground1) at (M.south) [ground,yshift=-1.5cm,xshift=-0.5cm,anchor=north] {};

        \draw [spring] (ground1.north) -- ($(M.south east)!(ground1.north)!(M.south west)$) node[midway,left=1.5mm,draw=none]{$k$};
        
        \node (ground2) at (M.south) [ground,yshift=-1.5cm,xshift=0.5cm,anchor=north] {};
        \draw (ground1.north west) -- (ground2.north east);
        \draw [damper] (ground2.north) -- ($(M.south east)!(ground2.north)!(M.south west)$) node[midway,right=2.0mm, draw=none]{$c$};

        \begin{scope}[node distance=0.1cm and 0.5cm]
            \node [above right=of M, draw=none, inner sep=0pt] (Fmid) {};
            \draw [ultra thick,-latex] (M.east) -| (Fmid.center) node[midway, draw =none, above right=0.1cm] {$F(t)$};
        \end{scope}

    \end{tikzpicture}
    \caption{Mass-spring-damper SDOF system.}
    \label{fig:sdof}
\end{figure}

The response of a \textit{Single-Degree-of-Freedom} (SDOF) mass-damper-spring system (Figure \ref{fig:sdof}) is considered in this simple case study. Here, the dynamical model can be defined by the following equation of motion,
\begin{equation}
    m\ddot{x}(t) + c\dot{x}(t) + kx(t) = F(t)
    \label{eq:SDOF_eq_motion}
\end{equation}
where $m$, $c$ and $k$ are the dynamic coefficients corresponding to the mass, damping and stiffness of the system, $x(t)$ is the response at a given time instance $t$, and $F(t)$ is the input force. The dots over the variables in equation (\ref{eq:SDOF_eq_motion}) denote the derivatives of the variable with respect to time. An impulse response, $h(t)$, was simulated here by solving equation~(\ref{eq:SDOF_eq_motion}) for a given selection of coefficients. In particular, the mass, damping and stiffness coefficients were given values of $m=1$, $c=20$, and $k=1\times10^{6}$, respectively.

Using the simulated target function, training samples were generated by the following,
\begin{equation}
    \mathbf{y}(t) = \mathbf{h}(t) + \epsilon
    \label{eq:training_samples}
\end{equation}
where $\epsilon$ corresponds to additive noise term, which follows a normal distribution $\epsilon \sim \mathcal{N}(0, \sigma^2)$. The noise is defined by a signal-to-noise ratio (SNR) equal to ten. Four training sample sets of varying sizes were created by selecting points at intervals of $16$, $12$, and $4$, whereby the input-training samples $t$ are uniform in $\left [ 0,0.3 \right ]$. Therefore, each training set ended up with sample sizes of $n=63$, $n=126$ and $n=251$, respectively.

\subsection{Predictors for the regression problem}
In this study, the regression problem is defined by following the kernel smoother representation,
\begin{equation}
    f(t_*) = \mathbf{k}(t_*)^T (\mathbf{K} + \sigma_n^2\mathbf{I})^{-1} \mathbf{y}
    \label{eq:dual_representation}
\end{equation}
where $\mathbf{k}(t) \in \mathbb{R}^N$ is a vector with elements $k_n = k(t_n,t)$, defined by the \textit{kernel function} $k(\cdot,\cdot)$, and evaluated for a new input $t_*$ with respect to all inputs in the training set. $\mathbf{K} \in \mathbb{R}^{N \times N}$ is the \textit{Gram matrix} with elements $K_{nm}=k(t_n,t_m)$, $\mathbf{I} \in \mathbb{R}^{N \times N}$ is the identity matrix and $\sigma_n^2 \geq 0$.

The ability of the learner to predict the correct output relies mostly on the choice of kernel function. In here, two different kernel functions are compared in the approximation of $\mathbf{h}(\mathbf{t})$. The first of these is the \textit{squared exponential}~\cite{Rasmussen2005} (SE), which has the form,
\begin{equation}
    k_{se}(t,t') = \sigma_{f_{se}}^2 \exp {\left ( -\frac{||t-t'||}{2l^2} \right )}
    \label{eq:se_kernel}
\end{equation}
where $\sigma_{f_{se}}$ and $l$ are the adjustable (hyper) parameters that control the characteristic form of the function. The second kernel function corresponds to that derived by Cross in~\cite{Cross2021}, whereby the physics of a linear SDOF system are embedded in the definition of the kernel function.
\begin{definition}[SDOF kernel function~\cite{Cross2021}]
    \textit{For a single degree-of-freedom linear oscillator with mass $m$, damping $c$ and stiffness $k$, the SDOF kernel function is defined as,}
    \begin{equation}
        k_{sdof}(t,t') = \frac{\sigma_{f_{sdof}}^2}{4m^2\zeta\omega_n^3} \exp (-\zeta\omega_n|t-t'|) \left [ \cos(\omega_d|t-t'|) + \frac{\zeta\omega_n}{\omega_d}\sin(\omega_d|t-t'|) \right ]
        \label{eq:sdof_kernel}
    \end{equation}
    \textit{where the natural frequency $\omega_n=\sqrt{k/m}$, the damping ratio $\zeta=c/2\sqrt{km}$, and the damped natural frequency $\omega_d=\omega_n\sqrt{1-\zeta^2}$.}
\end{definition}

Unlike the SE kernel, one may note that the SDOF kernel is defined by parameters that are readily interpretable. In~\cite{Cross2021}, it is shown that the SDOF kernel is better suited at predicting the response of the mass-damper-spring system to a random excitation. More interestingly, however, is its ability to allow the approximating function to extrapolate beyond the limits of the training set. This outcome is a consequence of embedding the relevant physics in the kernel function, and an aspect that models based solely on data lack, such as those in which the SE kernel is employed. Having a model that can extrapolate means better generalision. One should note that this claim only holds true for this particular scenario. Should the model present nonlinearities, for example, then it may then be unreasonable to expect good accuracy from using the SDOF kernel. Nonetheless, the aim pursued here is to formalise  generalisation when employing these kernel functions from the viewpoint of SLT.

\subsection{Model selection strategy}
For the model selection process, the SRM induction principle was employed. This approach required the estimation of the upper bound~\eqref{eq:regression_bound}, which in turn required the estimation of the VC dimension corresponding to each set of kernel smoothers. Unlike the case of using the parametric representation of~\eqref{eq:dual_representation}, the VC dimension of kernel smoothers cannot be determined easily. Some progress can be achieved when translating the kernel representation into an \textit{equivalent basis function expansion}~\cite{Cherkassky2007}, which is done by taking the eigendecomposition $\mathbf{K}=\sum_{i=1}^n \lambda_i \mathbf{u}_i \mathbf{u}_i^T$, where $\lambda_i$ is the $i$th eigenvalue and $\mathbf{u}_i$ is the corresponding eigenvector. Having the eigenvectors, the number of free parameters - or degrees of freedom - of a kernel can be defined as~\cite{Hastie1990},
\begin{equation}
    df = \sum_{i=1}^n \frac{\lambda_i}{\lambda_i + \sigma_n^2}
    \label{eq:kernel_dfs}
\end{equation}
and then taken as an estimate of the corresponding VC dimension. This definition roughly counts the number of prevalent eigenvalues, and therefore, if $\sum_{i=1}^n\lambda_i / (\lambda_i + \sigma_n^2) << 1$, then the contribution of the components in $\mathbf{y}$ along $\mathbf{u}_i$ are effectively eliminated~\cite{Rasmussen2005}.

As in~\cite{Cherkassky2007}, the general expression for the upper bound~\eqref{eq:regression_bound} can be reduced to,
\begin{equation}
    R(\theta) \leq \frac{1}{n}\sum_{i=1}^n(y_i - f(\mathrm{x}_i,\theta))^2 \left (1 - \sqrt{p-p\ln{p} + \frac{\ln{n}}{2n}} \right )_{+}^{-1}
    \label{eq:simple_bound}
\end{equation}
where the empirical risk is defined by the \textit{mean-squared-error} (MSE) and $p=h/n$. In this expression, the penalisation term is given by setting the constants $c=1$, $a_1=1$, $a_2=1$ and $\delta=4/\sqrt{n}$.

Now, given a kernel function, the model-selection structure can be defined in terms of the smoothing parameters $\theta$. For example, the length-scale parameter in~\eqref{eq:se_kernel} is known to strongly influence the rate at which the smoother varies, assuming the remaining parameters remain constant. A reduction in the length-scale produces functions that appear much more wiggly~\cite{Rasmussen2005}, leading to a slower decay in the corresponding eigenvalues~\cite{Hastie1990} (when sorted in descending order). Consequently, equation~\eqref{eq:kernel_dfs} would yield a higher estimate of the degrees of freedom, and thus, a higher estimate of the complexity (i.e. VC dimension). Some structure could then be defined on the set of SE kernel smoothers as,
\begin{equation}
    S_k = \left \{ f(k_{se}(t,t';l),\sigma_n),l \geq c_k \right \}, \quad \text{where}\ c_1 > c_2 > \dots
    \label{eq:se_structure}
\end{equation}
Each element in the structure is a subset of the next because the length-scale can take smaller values, thereby allowing for smoothers of higher complexity. Finally, the structure can be easily extended to include the effects of the remainder parameters on the complexity of the smoother.

This approach adheres to the SRM inductive principle in the sense that step ($1$) involves the selection of an element in~\eqref{eq:se_structure} (i.e.\ complexity estimation), followed by step ($2$) on determining the optimal parameters $\theta*$ that minimise the upper bound on the expected risk~\eqref{eq:simple_bound}. However, it should be noted that a structure can only be defined for one type of kernel function at a time. Two distinct structures were thus constructed. For each training set size, the fitting experiment was repeated $100$ times for a given realisation of the training set~\eqref{eq:training_samples}. At each iteration, the set $\theta$ corresponding to each kernel function was found by employing an exhaustive search over a range of values. In the case of the SE kernel, a set of possible values was defined for increasing values of the signal variance $\sigma_{f_{se}}$, and decreasing values of the length-scale $l$. As for the SDOF kernel, the set of possible values was only defined for the signal variance $\sigma_{f_{sdof}}$, under the assumption that the remaining parameters could be computed analytically from knowing the dynamic coefficients in advance. The noise parameter $\sigma_n$ was assumed to be known in all cases. The model-comparison strategy finally amounted to choosing the best model from the structure providing the minimum guaranteed risk~\eqref{eq:simple_bound}.

\subsection{Model selection strategy}
For the model selection process, the SRM induction principle was employed. This approach required the estimation of the upper bound~\eqref{eq:regression_bound}, which in turn required the estimation of the VC dimension corresponding to each set of kernel smoothers. Unlike the case of using the parametric representation of~\eqref{eq:dual_representation}, the VC dimension of kernel smoothers cannot be determined easily. Some progress can be achieved when translating the kernel representation into an \textit{equivalent basis function expansion}~\cite{Cherkassky2007}, which is done by taking the eigendecomposition $\mathbf{K}=\sum_{i=1}^n \lambda_i \mathbf{u}_i \mathbf{u}_i^T$, where $\lambda_i$ is the $i$th eigenvalue and $\mathbf{u}_i$ is the corresponding eigenvector. Having the eigenvectors, the number of free parameters - or degrees of freedom - of a kernel can be defined as~\cite{Hastie1990},
\begin{equation}
    df = \sum_{i=1}^n \frac{\lambda_i}{\lambda_i + \sigma_n^2}
    \label{eq:kernel_dfs}
\end{equation}
and then taken as an estimate of the corresponding VC dimension. This definition roughly counts the number of prevalent eigenvalues, and therefore, if $\sum_{i=1}^n\lambda_i / (\lambda_i + \sigma_n^2) << 1$, then the contribution of the components in $\mathbf{y}$ along $\mathbf{u}_i$ are effectively eliminated~\cite{Rasmussen2005}.

As in~\cite{Cherkassky2007}, the general expression for the upper bound~\eqref{eq:regression_bound} can be reduced to,
\begin{equation}
    R(\theta) \leq \frac{1}{n}\sum_{i=1}^n(y_i - f(\mathrm{x}_i,\theta))^2 \left (1 - \sqrt{p-p\ln{p} + \frac{\ln{n}}{2n}} \right )_{+}^{-1}
    \label{eq:simple_bound}
\end{equation}
where the empirical risk is defined by the \textit{mean-squared-error} (MSE) and $p=h/n$. In this expression, the penalisation term is given by setting the constants $c=1$, $a_1=1$, $a_2=1$ and $\delta=4/\sqrt{n}$.

Now, given a kernel function, the model-selection structure can be defined in terms of the smoothing parameters $\theta$. For example, the length-scale parameter in~\eqref{eq:se_kernel} is known to strongly influence the rate at which the smoother varies, assuming the remaining parameters remain constant. A reduction in the length-scale produces functions that appear much more wiggly~\cite{Rasmussen2005}, leading to a slower decay in the corresponding eigenvalues~\cite{Hastie1990} (when sorted in descending order). Consequently, equation~\eqref{eq:kernel_dfs} would yield a higher estimate of the degrees of freedom, and thus, a higher estimate of the complexity (i.e. VC dimension). Some structure could then be defined on the set of SE kernel smoothers as,
\begin{equation}
    S_k = \left \{ f(k_{se}(t,t';l),\sigma_n),l \geq c_k \right \}, \quad \text{where}\ c_1 > c_2 > \dots
    \label{eq:se_structure}
\end{equation}
Each element in the structure is a subset of the next because the length-scale can take smaller values, thereby allowing for smoothers of higher complexity. Finally, the structure can be easily extended to include the effects of the remainder parameters on the complexity of the smoother.

This approach adheres to the SRM inductive principle in the sense that step ($1$) involves the selection of an element in~\eqref{eq:se_structure} (i.e.\ complexity estimation), followed by step ($2$) on determining the optimal parameters $\theta*$ that minimise the upper bound on the expected risk~\eqref{eq:simple_bound}. However, it should be noted that a structure can only be defined for one type of kernel function at a time. Two distinct structures were thus constructed. For each training set size, the fitting experiment was repeated $100$ times for a given realisation of the training set~\eqref{eq:training_samples}. At each iteration, the set $\theta$ corresponding to each kernel function was found by employing an exhaustive search over a range of values. In the case of the SE kernel, a set of possible values was defined for increasing values of the signal variance $\sigma_{f_{se}}$, and decreasing values of the length-scale $l$. As for the SDOF kernel, the set of possible values was only defined for the signal variance $\sigma_{f_{sdof}}$, under the assumption that the remaining parameters could be computed analytically from knowing the dynamic coefficients in advance. The noise parameter $\sigma_n$ was assumed to be known in all cases. The model-comparison strategy finally amounted to choosing the best model from the structure providing the minimum guaranteed risk~\eqref{eq:simple_bound}.

\section{Results and discussion}

\begin{figure}[tp]
    \centering
    \begin{subfigure}[b]{0.32\textwidth}
        \centering
        \includegraphics[width=\textwidth]{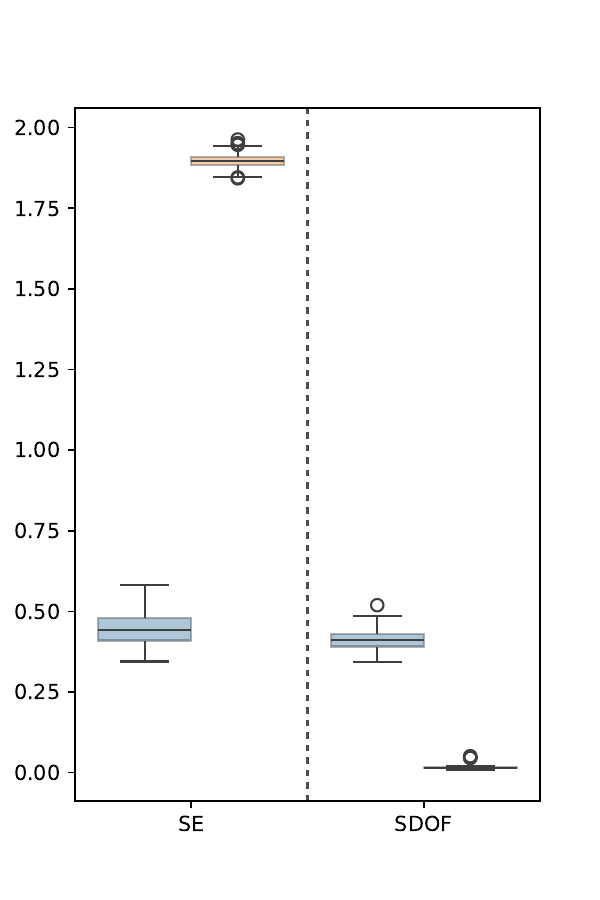}
        \caption{}
        \label{fig:sample_size_4}
    \end{subfigure}
    \begin{subfigure}[b]{0.32\textwidth}
        \centering
        \includegraphics[width=\textwidth]{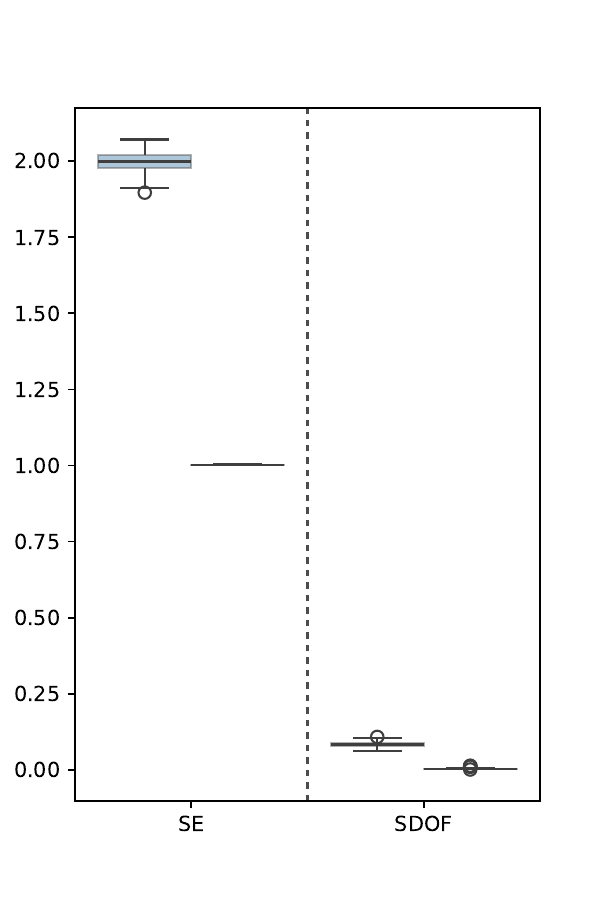}
        \caption{}
        \label{fig:sample_size_8}
    \end{subfigure}
    \begin{subfigure}[b]{0.32\textwidth}
        \centering
        \includegraphics[width=\textwidth]{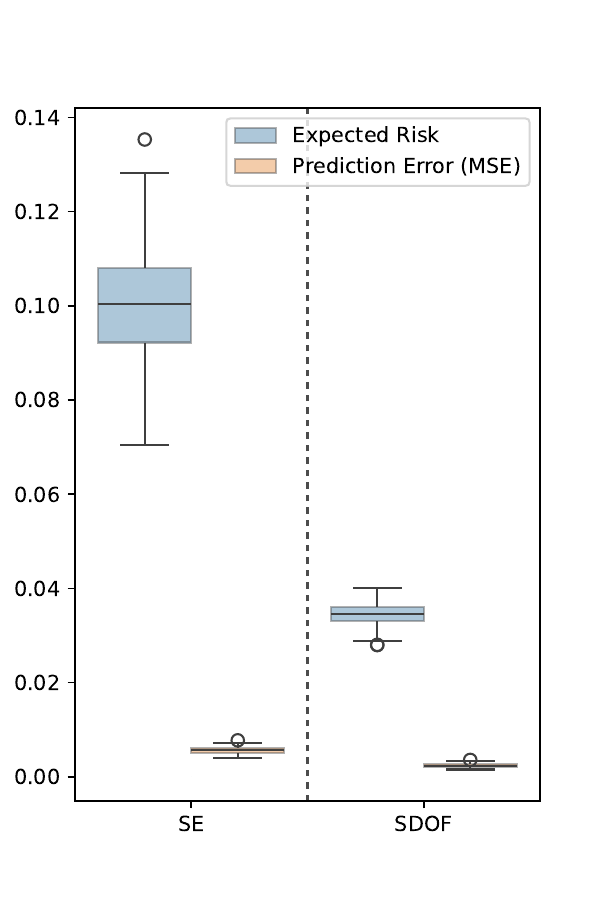}
        \caption{}
        \label{fig:sample_size_16}
    \end{subfigure}
    \caption{Expected risk and predition error (MSE) computed for the SE and SDOF kernel function, with training sets of sizes (a) $n=63$, (b) $n=126$, and (c) $n=251$.}
    \label{fig:boxplot_results}
\end{figure}

The results of the regression experiment are shown in Figure~\ref{fig:boxplot_results}. More specifically, for each sample size, the boxplots display the estimated upper bound on the expected risk, and the MSE between the true function and kernel-smoother predictions. Regardless of the sample size, the results show that expected risk was lower in all cases when choosing the SDOF kernel. The SRM inductive principle seems to indicate a preference for the SDOF model even when a handful of samples are available. However, one can note that such a conclusion is not so obvious when looking closer at the estimations obtained for the sample size $n=63$. In fact, the MSE is much higher than the estimated upper bound, which appears to contradict Theorem~\ref{th:regression_bound}. Upon closer inspection, a simple explanation to this outcome is explained by having established these bounds to hold with probability of at least $1-\delta$, where $\delta = 4/\sqrt{n}$. When having a small sample size, such as $n=63$, the resulting bounds hold with probability of at least $0.49$, and likely to yield bounds that may be too \say{optimistic}. In the case of having $n=126$ and $n=251$ as in the other two cases, the bounds hold with (higher) probabilities of at least $0.64$ and $0.75$, respectively. The intuition behind this trend is clear; that is, one can be more confident of the estimates of the risk when the sample size increases. If required, a low value of $\delta$ could be enforced to ensure the bounds hold with high probability. While bounds can be made more \say{pessimistic}, these could become too loose to provide any meaningful interpretation. Therefore, determining $\delta(n)$ as a function of the available number of samples is recommended~\cite{Cherkassky2007,Vapnik2006}.
		
\begin{figure}[tp]
    \centering
    \begin{subfigure}{0.80\textwidth}
        \centering
        \includegraphics[width=\textwidth]{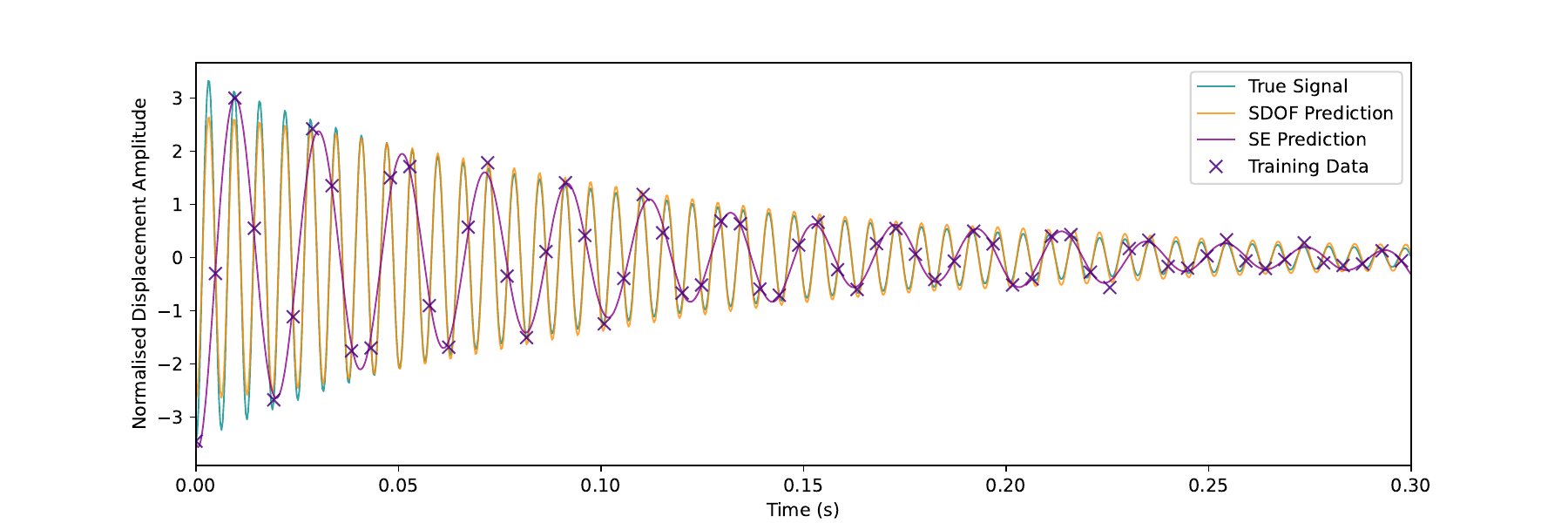}
        \caption{}
        \label{fig:pred_size_4}
    \end{subfigure}
    \begin{subfigure}{0.80\textwidth}
        \centering
        \includegraphics[width=\textwidth]{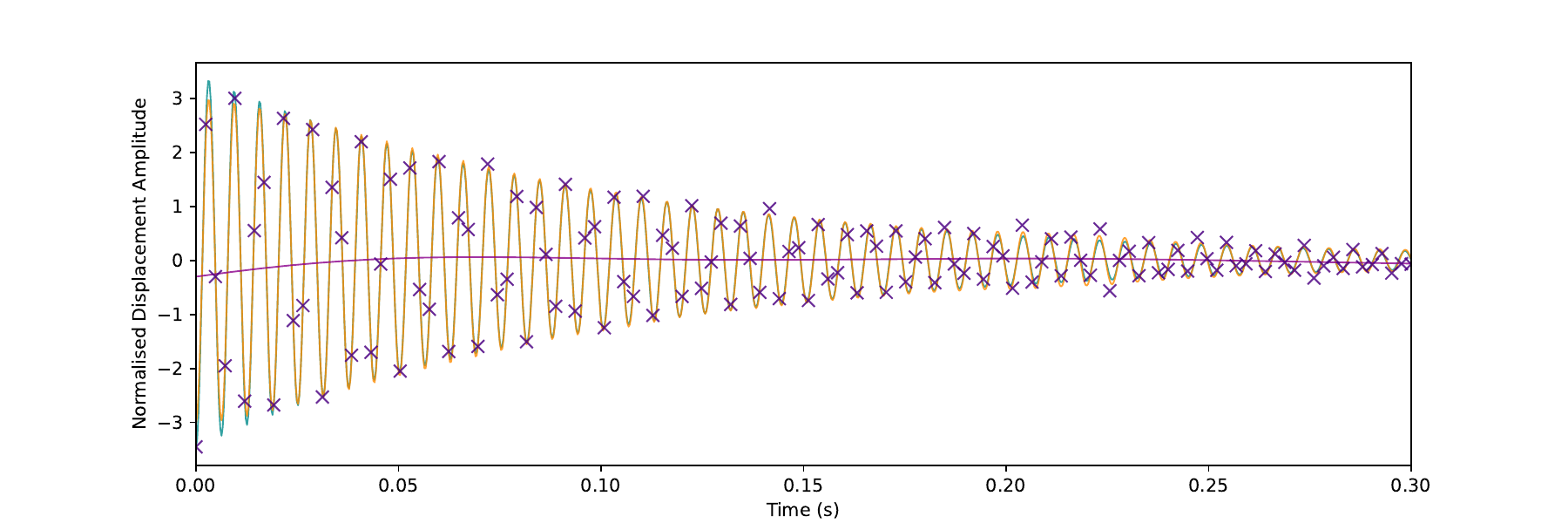}
        \caption{}
        \label{fig:pred_size_8}
    \end{subfigure}
    \begin{subfigure}{0.80\textwidth}
        \centering
        \includegraphics[width=\textwidth]{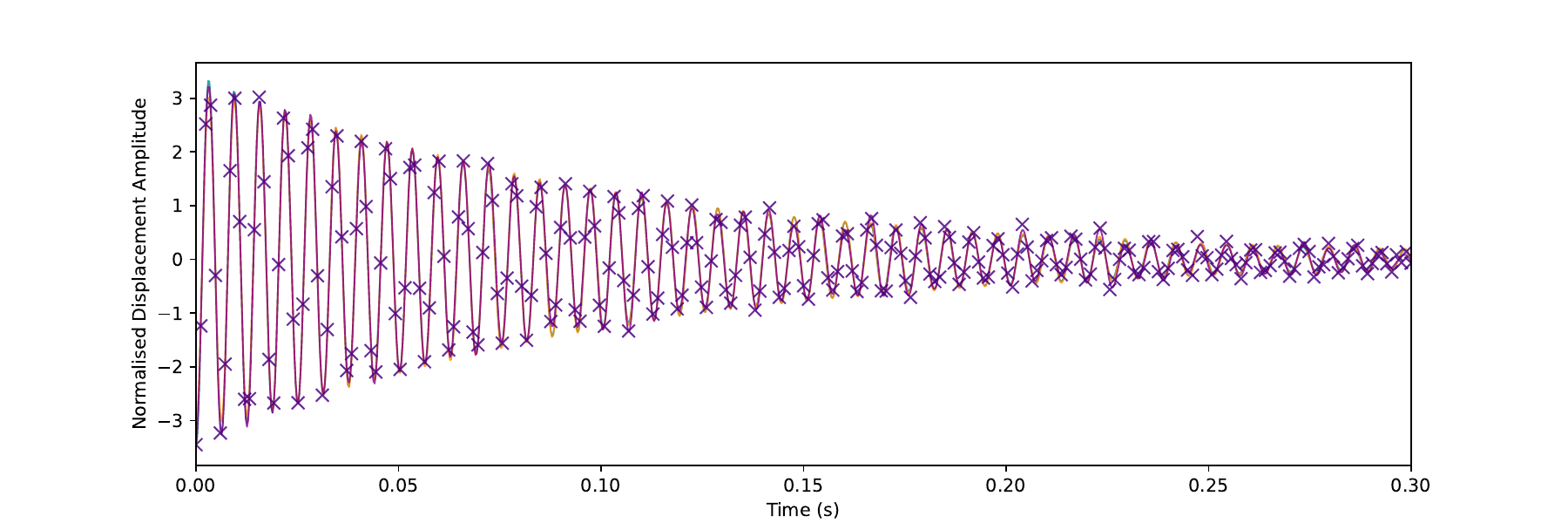}
        \caption{}
        \label{fig:pred_size_16}
    \end{subfigure}
    \caption{SE and SDOF kernel-smoother predictions inferred with training sets of sizes (a) $n=63$, (b) $n=126$, and (c) $n=251$. The true underlying signal was included for comparison.}
    \label{fig:pred_results}
\end{figure}

The kernel-smoother predictions of the true function can be visualised in Figure~\ref{fig:pred_results}. Each figure corresponds to the different training sets used in the model selection process, showing the true signal, predictions from employing each kernel function, and the training data realisation from an arbitrary iteration. As evidenced by the results in Figure~\ref{fig:boxplot_results}, the SDOF kernel smoother tends to generalise better, even when the sample size is small. It is not until $n=251$ that the SE kernel smoother learns the true underlying function from the data. The increased sample size appears to have relaxed the penalisation term enough to permit higher complexities.

Perhaps of more interest is how the increase in sample size seemed to not only minimise the expected risk, but also bring their estimates closer to the same value regardless of the kernel function used. If enough data samples are thus available, then one may argue the necessity for developing informed kernels (or any approximating function for that matter) to be a superfluous exercise. There are several reason why this claim may not be entirely true. While tighter bounds are, indeed, possible with large datasets, it may not be computationally efficient to have large training sets. Therefore, a model that can represent the system as well as another that requires significantly more data to train should be favoured in the model selection process. Additionally, the availability of labelled data can be very limited in certain applications. For example, in SHM, it is more often the case than not that labelled data are expensive to gather, and the efficacy of statistical models is likely to rely on a reduced dataset. SLT embodies the notion that a lack of data should be compensated with knowledge for better inference, and the results shown here are but a simple demonstration of this notion in SHM.

\begin{figure}[tp]
    \centering
    \includegraphics[width=0.66\textwidth]{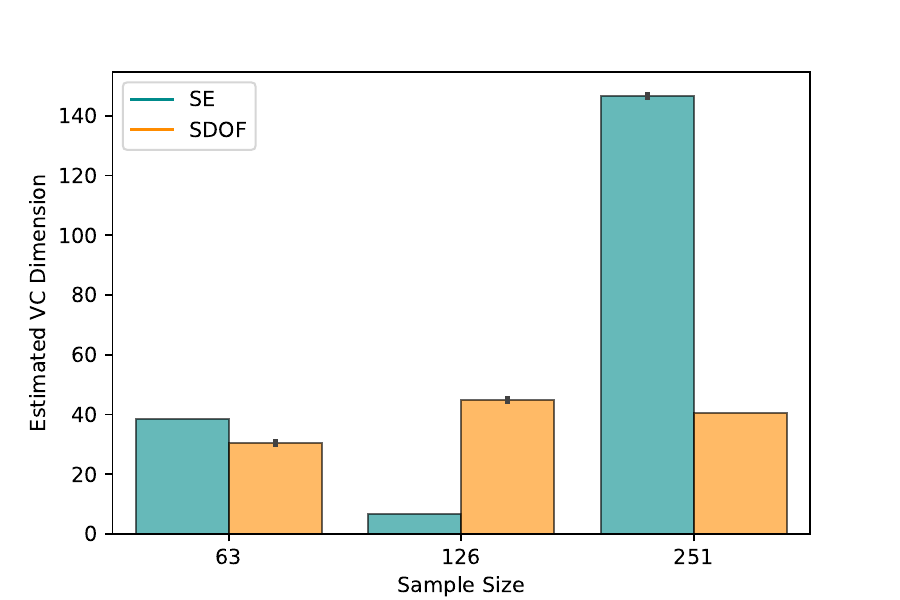}
    \caption{Estimated complexities corresponding to the SE and SDOF kernel with training sets of sizes (a) $n=63$, (b) $n=126$, and (c) $n=251$.}
    \label{fig:complexities}
\end{figure}

One final highlight that might be worthy of discussion is on the flexible nature exhibited by the SE-kernel function. Visually, as shown in Figure~\ref{fig:pred_results}, the SE smoothing functions displayed varied complexities in all three cases. The rate at which these predictors vary appear to somewhat agree with the respective VC dimensions shown in Figure~\ref{fig:complexities}. These estimates are the optimal complexities provided by the SRM inductive principle. As noted earlier, the complexity of the SE kernel smoother for $n=63$ was relatively high because of having too optimistic a bound. For the other two cases, however, the models guaranteed better generalisation, even if it may not seem like it (visually) in Figure~\ref{fig:pred_size_8}. Computing the empirical risk alone for $n=126$ would have caused the SE-kernel smoother to overfit the data. In such cases the bound exploded to infinity because of the relatively high ratio of complexity to sample size. Conversely, the high VC dimension displayed in Figure~\ref{fig:complexities} for $n=251$ was only possible because of having enough sample points to justify such increase in complexity. Unlike the SE kernel function, one may note that the VC dimension estimated from using the SDOF kernel function remained somewhat consistent for all sample sizes.

\section{Conclusions}

In the present work, the generalisation of kernel smoothers was defined from the viewpoint of SLT. In particular, the model-selection process was simulated in the context of a simple linear oscillator subjected to an impulse response. By employing the SRM inductive principle, the model-selection process involved the elicitation of a kernel smoother that could generalise best when given a reduced dataset. The two kernel functions considered in the experiment differ in their means to characterise neighboring similarities in the data; namely, the SE kernel was compared against the SDOF kernel, with the latter deriving from embedding the physics of the system under consideration. Even though the SDOF kernel function was expected to be a much more suitable model for this application, a mathematical framework to justify the choice of an informed kernel had not been established in this context. The bounds found in SLT offer the possibility to pursue a rigorous and systematic approach to justify the use of an informed kernel over another that is solely driven by data.
		
However, there are several shortcomings that require further attention to formally validate the results presented in this paper. The first and foremost, is that it is not clear how accurate the number of degrees of freedom of a kernel smoother estimates the VC dimension. Estimating the complexity for nonparametric sets of functions is a subject of ongoing research. The second issue relates to the structure design in the model-selection process. In the study above, to distinct structures were developed, followed by the selection of the minimal risk given by the optimal elements of each. This criterion was based on the idea that each structure may correspond to separate subsections in a vast space of all possible approximating functions. The subsection relating to the SDOF kernel was expected to enclose the true function, and therefore, the true guaranteed risk. Meanwhile, the subsection spanned by the SE kernel function required the extension to parent elements in the structure to approximate closer to the true function. Consequently, the SRM inductive principle accounted for two risk minima corresponding to each structure. Favouring the lowest of the two was then the natural step in selecting the optimal model. The issue here is that the validity of this hypothesis is not one formally addressed in SLT. It may be necessary to refer to the methods found in \textit{decision theory} for this matter. Finally, the last shortcoming that should be acknowledge is on the fact that the dynamic coefficients were provided in the SDOF kernel function. If these coefficients were to be unknown, then their inclusion in the structural minimisation process would be necessary. This issue and the aforementioned ones are certainly worthy of further research, and will be addressed in future work.

\section{Acknowledgments}

The authors of this paper gratefully acknowledge the support of the UK Alan Turing Institute via funding of the Turing Research and Innovation Cluster on Digital Twins. For the purpose of open access, the authors has applied a Creative Commons Attribution (CC BY) licence to any Author Accepted Manuscript version arising.

\bibliographystyle{unsrt}
\bibliography{refs}

\end{document}